\documentclass[conference]{IEEEtran}
\usepackage{cite}

\ifCLASSINFOpdf
  \usepackage[pdftex]{graphicx}
  \usepackage{subfigure}
  \usepackage{multicol}
  \usepackage{multirow}
  \usepackage{booktabs}
\else
\fi
\usepackage{amsmath}
\usepackage{array}
\begin{document}
%
\title{Online Signature Verification using Recurrent Neural Network and Length-normalized Path Signature Descriptor}

\author{\IEEEauthorblockN{Songxuan Lai, Lianwen Jin, Weixin Yang}
\IEEEauthorblockA{School of Electronic and Information Engineering\\
South China University of Technology\\
Guangzhou, China\\
Email: lai.sx@mail.scut.edu.cn, eelwjin@scut.edu.cn}
}


%


\maketitle

\begin{abstract}
Inspired by the great success of recurrent neural networks (RNNs) in sequential modeling, we introduce a novel RNN system to improve the performance of online signature verification. The training objective is to directly minimize intra-class variations and to push the distances between skilled forgeries and genuine samples above a given threshold. By back-propagating the training signals, our RNN network produced discriminative features with desired metrics. Additionally, we propose a novel descriptor, called the length-normalized path signature (LNPS), and apply it to online signature verification. LNPS has interesting properties, such as scale invariance and rotation invariance after linear combination, and shows promising results in online signature verification. Experiments on the publicly available SVC-2004 dataset yielded state-of-the-art performance of 2.37\% equal error rate (EER).
\end{abstract}


%
\IEEEpeerreviewmaketitle

\section{Introduction}
Signature verification has been an active research area because of the long-term and widespread use of signatures for personal authentication, however it remains a challenging task owing to large intra-class variations and skilled forgeries. Signature verification can be categorized as online or offline, depending on the signature acquisition method. Traditionally, online signature verification systems perform better than offline systems because more dimensions of information are available.

This paper deals with online signature verification. Most previous works in online signature verification use Dynamic Time Warping (DTW \cite{Rabiner1993Fundamentals}) \cite{Kholmatov2005Identity}\cite{Nanni2010Combining}\cite{Sharma2016An} or Hidden Markov Models (HMMs \cite{Rabiner1993Fundamentals}) \cite{Fierrez2005An}\cite{Fierrez2007HMM}\cite{Van2007On}. DTW is an effective template-based method for online signature verification in which only small amounts of data is available. HMMs, with Gaussian mixture models (GMMs) as hidden states, can be regarded as a soft version of DTW and outperforms DTW when enough training signatures are available \cite{Fierrez2005An}. Although progress has been made on both DTW and HMM systems, the performance of online signature verification systems is, still, a concern compared to other biometrics such as fingerprint and iris scans.

Inspired by the great success of recurrent neural network (RNNs) in sequential modeling \cite{Graves2012Supervised}\cite{Gregor2015DRAW}\cite{Donahue2016Long}, we introduce a novel RNN system to further improve online signature verification. Because training an RNN requires a relatively large dataset, we combine several existing online signature datasets to jointly train our system. The training objective is to directly minimize intra-class variations and to increase the distances of skilled forgeries from genuine samples past a given threshold, which is achieved by triplet loss \cite{hoffer2015deep} and center loss \cite{wen2016discriminative}. By back-propagating the training signals, the RNN network produces discriminative features with desired metrics. In addition, a new descriptor, called the length-normalized path signature (LNPS), is proposed and introduced for online signature verification. The LNPS descriptor is scale invariant and, after some linear operation, rotation invariant, and encodes contextual information from a window sliding over the signature. The overall architecture of the proposed online signature verification system is presented in Fig. 1.

\begin{figure*}
  \centering
  \includegraphics[width=6in]{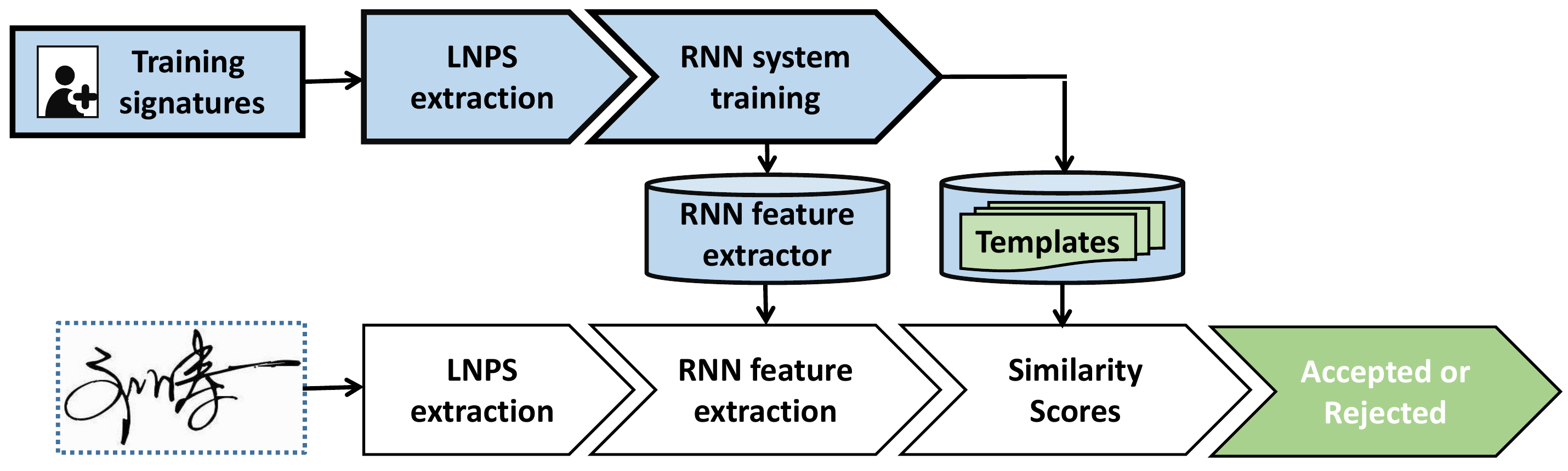}
  \caption{Architecture of our online signature verification system.}\label{system}
\end{figure*}

The remainder of this paper is organized as follows. Section II introduces the LNPS descriptor and its properties. Section III describes the architecture of our proposed RNN system. Section IV describes the decision making method. Experimental results are given and analyzed in section V. Section VI outlines some conclusions of our study.

\section{Feature extraction using length-normalized path signature}

\subsection{Pre-processing}
Many preprocessing methods have been used in online signature verification systems, including smoothing paths, rotation and scale normalization, connecting consecutive strokes with virtual strokes. Because our proposed length-normalized path signature (LNPS) descriptor is scale invariant (and rotation invariant after some linear transformation), we do not apply any preprocessing to the signatures.

\subsection{Length-normalized path signature}
In mathematics, the path signature is a collection of iterated integrals of a path \cite{chevyrev2016primer}. Let \emph{X}:[0, \emph{T}]$\rightarrow$\emph{$R^d$} denote a continuous path of bounded variation, mapping from time interval [0, \emph{T}] to space \emph{$R^d$}. Then the \emph{k}$^{th}$ level iterated integral of path \emph{X} is
\begin{equation}\label{pathIntegral}
  I^k(X)=\int_{0<t_1<...<t_k<T}1dX_{t_1}\otimes\ldots\otimes dX_{t_k},
\end{equation}
where $\otimes$ represents the tensor product. By convention, $I^0$ is the number one. The path signature truncated at level \emph{m} is the collection of iterated integrals up to level \emph{m} of path \emph{X}:
\begin{equation}\label{signature}
  S(X)|_m=\left[
            \begin{array}{ccccc}
              1 & I^1(X) & I^2(X) & ... & I^m(X) \\
            \end{array}
          \right]^T.
\end{equation}
If \emph{X} is sampled and approximated by a set of discrete points, then \emph{$S(X)|_m$} can be approximated by using simple tensor algebra \cite{graham2013sparse}. The path signature is an effective feature representation and has achieved significant success in handwriting recognition \cite{graham2013sparse}\cite{yang2015chinese}\cite{yang2016deepwriterid}.

An online signature is exactly such a path mapping from [0, \emph{T}] to \emph{$R^d$}, where \emph{d} is the dimension of available time functions during data acquisition. In this work, we only use the x-y coordinates, hence \emph{d}=2. By sliding over an online signature $D$, we can extract path signature within the sliding window at every position. Assume the window size is $W=2\times ws+1$, then the data sequence within the window at position \emph{n} is
\begin{equation}\label{sigWin}
  d(n)=\left[
            \begin{array}{ccccccc}
                x_{n-ws} & ... & x_n & ... & x_{n+ws} \\
                y_{n-ws} & ... & y_n & ... & y_{n+ws} \\
            \end{array}
            \right].
\end{equation}
Path signature is calculated from $d(n)$, resulting in a feature vector $S(d(n))|_m$ that encodes contextual information. Thus, the online signature $D$ can be represented as a time sequence:
\begin{equation}\label{sigSeq}
  D = \left[ \begin{array}{cccc}
        S(d(1))|_m &
        S(d(2))|_m &
        ... &
        S(d(N))|_m
      \end{array} \right],
\end{equation}
where $N$ is the number of points of $D$. However, the values of $S(X)|_m$ vary with the scale of path X, and different levels of iterated integrals $I^k$ usually have values of different orders of magnitude. To address this issue, we use the length of $X$ to normalize $S(X)|_m$ and propose the novel LNPS descriptor. Specifically, first calculate the length of $X$, denoted as $L$. Then length $L$ is used to normalize $S(X)|_m$ in the following way, resulting in the LNPS:
\begin{equation}\label{LNPS}
  S(X)|_m^{LN}=\left[
            \begin{array}{ccccc}
              1 & \frac{I^1(X)}{L(X)} & \frac{I^2(X)}{L^2(X)} & ... & \frac{I^m(X)}{L^m(X)} \\
            \end{array}
          \right]^T.
\end{equation}
Note that in the above equation, $L^k$ means the \emph{k}$^{th}$ power of $L$, which differs from $I^k$. Therefore, $D$ is represented as
\begin{equation}\label{sigSeqLNPS}
  D = \left[ \begin{array}{cccc}
        S(d(1))|_m^{LN} &
        S(d(2))|_m^{LN} &
        ... &
        S(d(N))|_m^{LN}
      \end{array} \right],
\end{equation}
which is then further z-normed channel-wise.

The resulting representation of $D$ has the following property:
\subsubsection{Scale invariance}
This can be easily derived from the definition in Eq. (1) and (5). Fig. 2(a) illustrates $\frac{I^1}{L}$, which is scale invariant.
\subsubsection{Rotation invariance after some linear operation}
Rotation invariants of two-dimensional paths has been studied in \cite{diehl2013rotation}, in which it was found that some linear combinations of iterated integrals are invariant to path rotation. For example, $I^2$ is a four-dimensional vector, in which
\begin{equation}\label{RotInv1}
  I^2(X)[2] = \int_{0<t_1<t_2<T}dX_{t_1}^{2}dX_{t_2}^{1},
\end{equation}
\begin{equation}\label{RotInv2}
  I^2(X)[3] = \int_{0<t_1<t_2<T}dX_{t_1}^{1}dX_{t_2}^{2}.
\end{equation}
Then according to the Green's theorem \cite{rudin1964principles}, $\frac{I^2(X)[3]-I^2(X)[2]}{2}$ is the area swept out by the closed path $X$, illustrated in Fig. 2(b), and therefore is invariant to the rotation of $X$. Other rotation invariants can be explicitly written out by linear combinations, but we leave it for the RNN network. If the rotation invariants are useful for the task, back-propagation (BP) \cite{williams1986learning} should be able to find them.
\subsubsection{Step over the stroke discontinuities}
In some previous work, virtual strokes are sometimes added to connect consecutive strokes. By using the normalization method in Eq. (6), feature transilience caused by stroke discontinuities is overcome.

The dimension of LNPS depends on the truncated level $m$, which is an important parameter in this work. Generally, higher levels of LNPS would characterize more detailed path information \cite{graham2013sparse}\cite{yang2015chinese}\cite{yang2016deepwriterid}.


\begin{figure}[tb]
  \centering
  \subfigure[A scale invariant feature from LNPS.]{\includegraphics[width=1.5in]{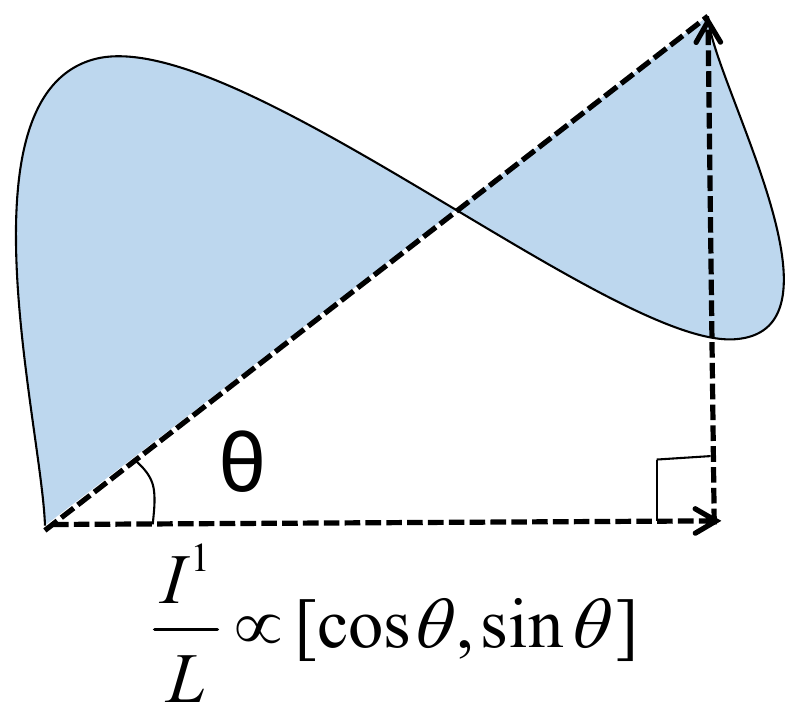}}
  \hspace{0.3in}
  \subfigure[A rotation invariant feature.]{\includegraphics[width=1.5in]{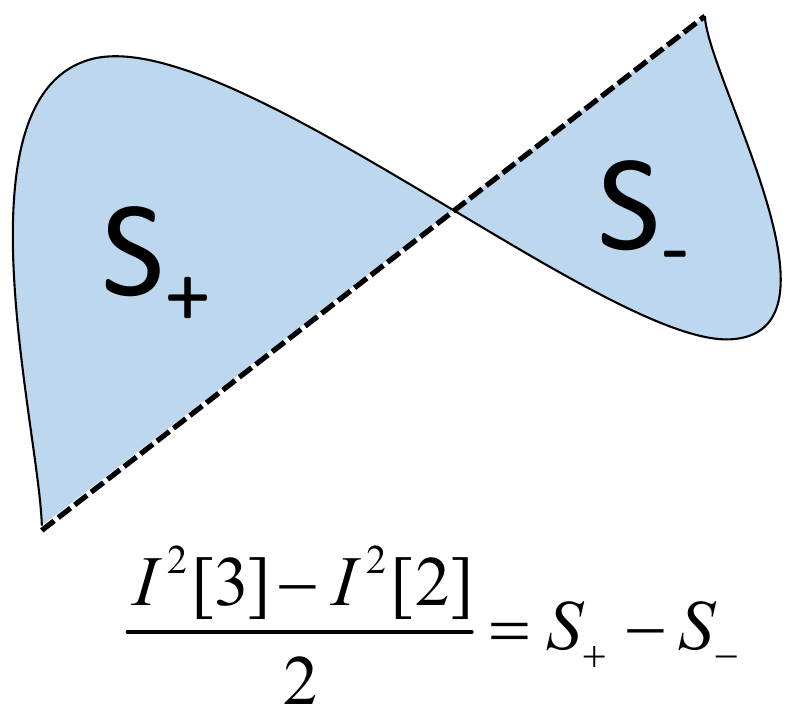}}
  \caption{Illustration of the property of the proposed LNPS.}\label{Area}
\end{figure}

\begin{figure*}[tb]
  \centering
  \includegraphics[width=6in]{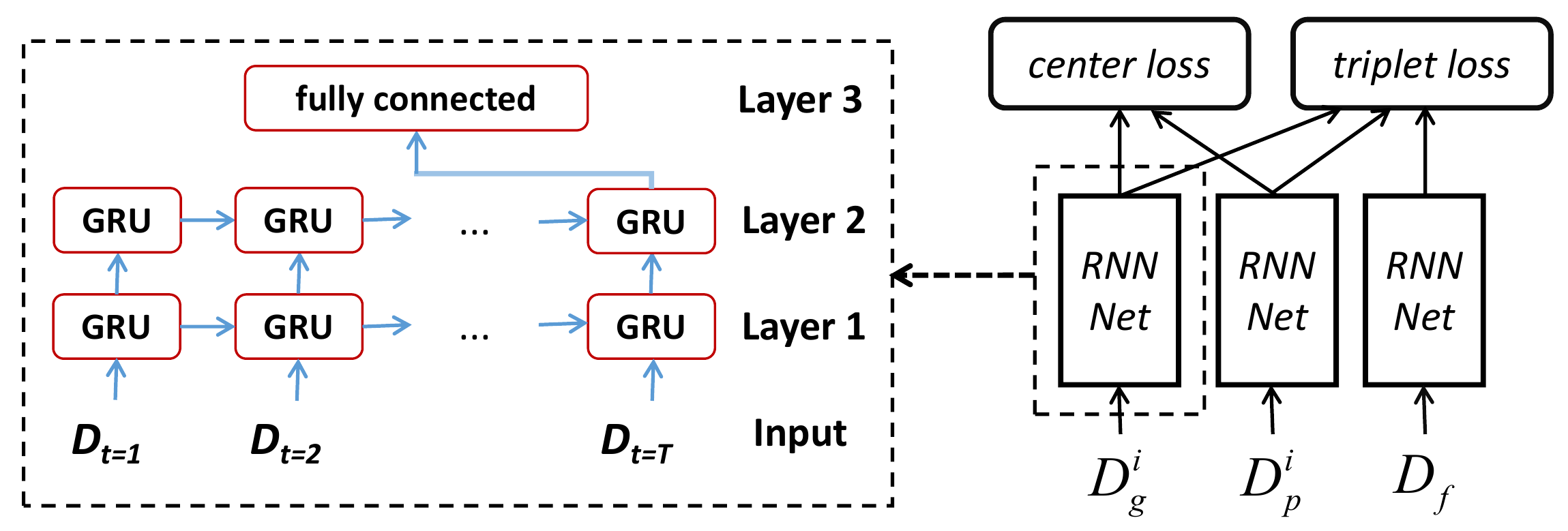}
  \caption{Our way of training the RNN system for online signature verification.}\label{RNN}
\end{figure*}

\section{RNN system with metric learning}
Recurrent architectures such as LSTM \cite{hochreiter1997long} and GRU \cite{cho2014properties} have exhibited state-of-the-art performance on many complex sequential problems \cite{Graves2012Supervised}\cite{Gregor2015DRAW}\cite{Donahue2016Long}. However, RNN's application to online signature verification is hindered by limited amounts of data in this task. To address this problem, we combine two online signature datasets, SVC-2004 \cite{yeung2004svc2004} and MCYT-100 \cite{ortega2003mcyt}, for joint training. Concretely, we train on both SVC-2004 and MCYT-100 and test on SVC-2004. Although these two datasets are written by different clients under different environments, we find that by joint training, knowledge learned from one dataset can transfer to the other dataset.

\subsection{Network architecture}
In this work, we use the gated recurrent unit (GRU) in our RNN network, as it has a simpler architecture than LSTM while exhibiting similar performance \cite{chung2014empirical}. The GRU controls the information flow inside the unit using reset gate $r_t$ and update gate $z_t$ :
\begin{equation}\label{reset}
  r_t = sigm(W_r\times x_t+U_r\times y_{t-1} + b_r),
\end{equation}
\begin{equation}\label{update}
  z_t = sigm(W_z\times x_t+U_z\times y_{t-1} + b_z),
\end{equation}
\begin{equation}\label{act}
  \widetilde{y_t} = tanh(W\times x_t+U\times (r_t\odot y_{t-1}) + b),
\end{equation}
\begin{equation}\label{renew}
  y_t = z_t\odot y_{t-1}+(1-z_t)\odot \widetilde{y_t},
\end{equation}
where $y_t$ is the GRU activation function and $x_t$ is the input at time step $t$. $W_r$, $W_z$, $W$ are the recurrent weights and $b_r$, $b_z$, $b$ are the biases. $sigm(\cdot)$ is the sigmoid function and $tanh(\cdot)$ is the tanh function.

We use the following simple architecture: Input-128GRU-128GRU-64FC, as shown on the left side of Fig. 3. A simple network is easy to optimize, thus suitable for the task of online signature verification. The output is a fixed-length 64-dimensional feature vector, a highly nonlinear function $G(\cdot)$ of the input.

\subsection{Loss functions}
Proper loss functions should be defined to train the GRU network with the BP algorithm. Online signature verification can be properly defined as a metric learning problem; therefore, loss functions in metric learning can be introduced. Let $D_g^{i}$ denote a genuine signature from client $i$, $D_p^{i}$ denote another genuine signature that also comes from client $i$, $D_f$ denote either a skilled or random forgery. \{$D_g^{i}$, $D_p^{i}$, $D_f$\} is called a triplet. To discriminate the genuine signatures from the forgeries, we should minimize the following loss function, defined on all triplets:
\begin{equation}\label{triplet}
  L_{t}=\sum max\{\|G(D_g^{i})-G(D_p^{i})\|-\|G(D_g^{i})-G(D_f)\|+C,0\}.
\end{equation}
$L_{t}$ is called the triplet loss \cite{hoffer2015deep}, where C is a threshold controlling the distance between $\|G(D_g^{i})-G(D_p^{i})\|$ and $\|G(D_g^{i})-G(D_f)\|$. Recall that $G(\cdot)$ returns the a 64-dimensional feature vector. By minimizing $L_{t}$, $G(\cdot)$ will be automatically adjusted by BP and produce meaningful feature vectors. In addition, center loss \cite{wen2016discriminative} is introduced to the minimize intra-class variations. Let $G^{i}$ denote the mean feature vector of client $i$, then
\begin{equation}\label{center}
  L_{c}=\sum \|G(D_g^{i})-G^{i}\|+\|G(D_p^{i})-G^{i}\|.
\end{equation}
$L_{c}$ is a summation over all triplets, as is $L_{t}$. Weight decay of the GRU network's fully connected layer ( i.e., layer 3 in Fig. 3) is also added to avoid feature scaling.

The overall loss function is therefore
\begin{equation}\label{loss}
  L=L_{t}+\lambda_{c}\times L_{c}+\lambda_{decay}\times L_{decay}.
\end{equation}
In our work, we set $C$=1, $\lambda_{c}$=0.5 and $\lambda_{decay}$=0.0001, and use the Euclidean distance for $\|\cdot\|$ in Eq. (13) and (14). Our way of training the RNN system (i.e., the RNN feature extractor) is presented in Fig. 3.
\subsection{Network training}
We use Adamax \cite{kingma2014adam} to train our RNN system. Adamax is a variant of Adam \cite{kingma2014adam} and has self-adaptive gradients. The initial learning rate is 0.01 and gradients are clipped between $[-1, 1]$. The system is trained for 400 epochs.

\section{Decision making}
Given a set of genuine signatures from client $i$ as templates, denoted as \{$D_1^{i}, D_2^{i}, ..., D_N^{i}$\}, and a test signature denoted as $D_{test}$, we should decide whether $D_{test}$ is a genuine signature from $i$ based on \{$D_1^{i}, D_2^{i}, ..., D_N^{i}$\}. In this work, we follow the simple decision method in \cite{Sharma2016An}. Denote the distances between $D_{test}$ and \{$D_1^{i}, D_2^{i}, ..., D_N^{i}$\} as \{$d_{1,test}^{i}, d_{2,test}^{i},..., d_{N,test}^{i}$\}, and the pairwise distances within \{$D_1^{i}, D_2^{i}, ..., D_N^{i}$\} as \{$d_{1,2}^{i}, d_{1,3}^{i}, ..., d_{1,N}^{i}, d_{2,3}^{i}, ..., d_{N-1,N}^{i}$\}. Then we have:
\begin{equation}\label{score}
  score_{test}^{i} = \frac{N-1}{2}\sum_{n=1}^{N}d_{n,test}^{i}/\sum_{n=1}^{N}\sum_{m>n}d_{n,m}^{i}.
\end{equation}
If $score_{test}^{i}$ is less than a (user-independent) global threshold, then $D_{test}$ is considered a genuine signature from client $i$. In the RNN system, distance $d$ between two samples $D_1$ and $D_2$ is computed as $\|G(D_1)-G(D_2)\|$.

\section{Experiments}
\subsection{Datasets}
The experiments were conducted on the publicly available SVC-2004 task2 \cite{yeung2004svc2004} and MCYT-100 \cite{ortega2003mcyt} datasets. Regarding the SVC-2004 dataset, signatures were collected from a WACOM Intuos tablet with 20 genuine samples and 20 skilled forgeries for each of 40 total individuals. The MCYT-100 dataset comprises signatures from 100 individuals, with 25 genuine samples and 25 forgeries for each individual. The signatures were collected using a WACOM Intuos A6 USB pen tablet. Our system only uses the x-y coordinates.

\subsection{Effectiveness of LNPS}
To validate the effectiveness of the proposed LNPS descriptor, a basic DTW enhanced by VQ \cite{Sharma2016An} was tested on the SVC-2004. For each client, five genuine signatures were randomly selected from the first 10 genuine samples and all remaining signatures were used as the test set. Each experiment was repeated for 10 times and the average skilled forgery equal error rate (EER) is reported. Note that each LNPS level was evaluated individually, i.e., $\frac{I^k}{L^k}$ was evaluated for each k. The rotation invariants up to level 4, denoted as LNPS$_{RI}$, were also evaluated. Decision making follows Section IV, with distance $d$ computed as the DTW score. We varied the sliding window size $W$ and LNPS level $k$ to evaluate their effects on performance. The results are presented in Table I.
\begin{table}[tb]
\renewcommand{\arraystretch}{1.2}
\caption{EER (\%) of the DTW system using LNPS descriptor, on SVC-2004 dataset.}
\label{DTW}
\centering
\fontsize{10}{12}\selectfont
\begin{tabular}{cccccc}
\hline
\multirow{2}{*}{LNPS}&
\multicolumn{5}{c}{Sliding window size}\cr
\cmidrule(lr){2-6}& $W$=7& $W$=9& $W$=11& $W$=13& $W$=15\cr
\midrule
$k$=1&8.70&8.12&7.91&7.99&8.29\\
$k$=2&6.64&5.92&5.67&5.44&5.94\\
$k$=3&6.39&5.16&5.56&5.36&5.59\\
$k$=4&5.51&5.10&5.35&4.98&5.34\\
\hline
LNPS$_{RI}$&8.57&6.23&5.26&5.47&5.60\\
\bottomrule
\end{tabular}
\end{table}

Table I shows that a higher LNPS level yields better performance, because it encodes more detailed path information. A moderate window size (9 to 13 points, about 100 ms at a 100-Hz sampling rate) is generally better than smaller or larger windows, because it encodes an appropriate range of contextual information. Some of the results are better than the SVC-2004 competition winner (5.50\% EER for 40 users), but are not comparable with state-of-art algorithms \cite{Nanni2010Combining}\cite{Sharma2016An}. However, our aim is not to tune the DTW system for optimal performance, but to provide an experimental proof of the LNPS's effectiveness in online signature verification.

\subsection{Evaluation of the proposed RNN system}
To train the RNN system with triplet loss, we need negative samples (i.e., forgeries) for generating triplets. Hence, in addition to genuine samples, an equal number of forgeries were also randomly selected from the skilled forgery samples. $N$ genuine and skilled forgery samples were randomly selected for each client, and all remaining samples were used as the test set.

First, experiments were conducted on SVC-2004 to evaluate the effects of sample number $N$, sliding window size $W$, and LNPS level $m$ (in Eq. (6)) on the RNN system. For joint training, the entire MCYT-100 dataset (100 extra clients) was added to the training set and helped to optimize the network. Each experiment was repeated for five times; the average skilled forgery EER is reported in Table II, from which we can see that—contrary to the DTW system—LNPS with level $m>$2 does not achieve better performance. This may be because the GRU is already good at sequential modeling, hence, extra input dimensions do not help, but instead lead to optimization difficulty. Fig. 4 shows the test scores in one experiment in which $W=9$, $m=2$ and $N=10$. Note that some scores are higher than 80 and thus not shown in the figure. When using commonly used $\triangle x$, $\triangle y$ as features (this is a special case of $I^1$ with a sliding window of size 2), the results are shown in Table III; again, these results validate the effectiveness of LNPS.

\begin{figure}
  \centering
  \includegraphics[width=3.5in]{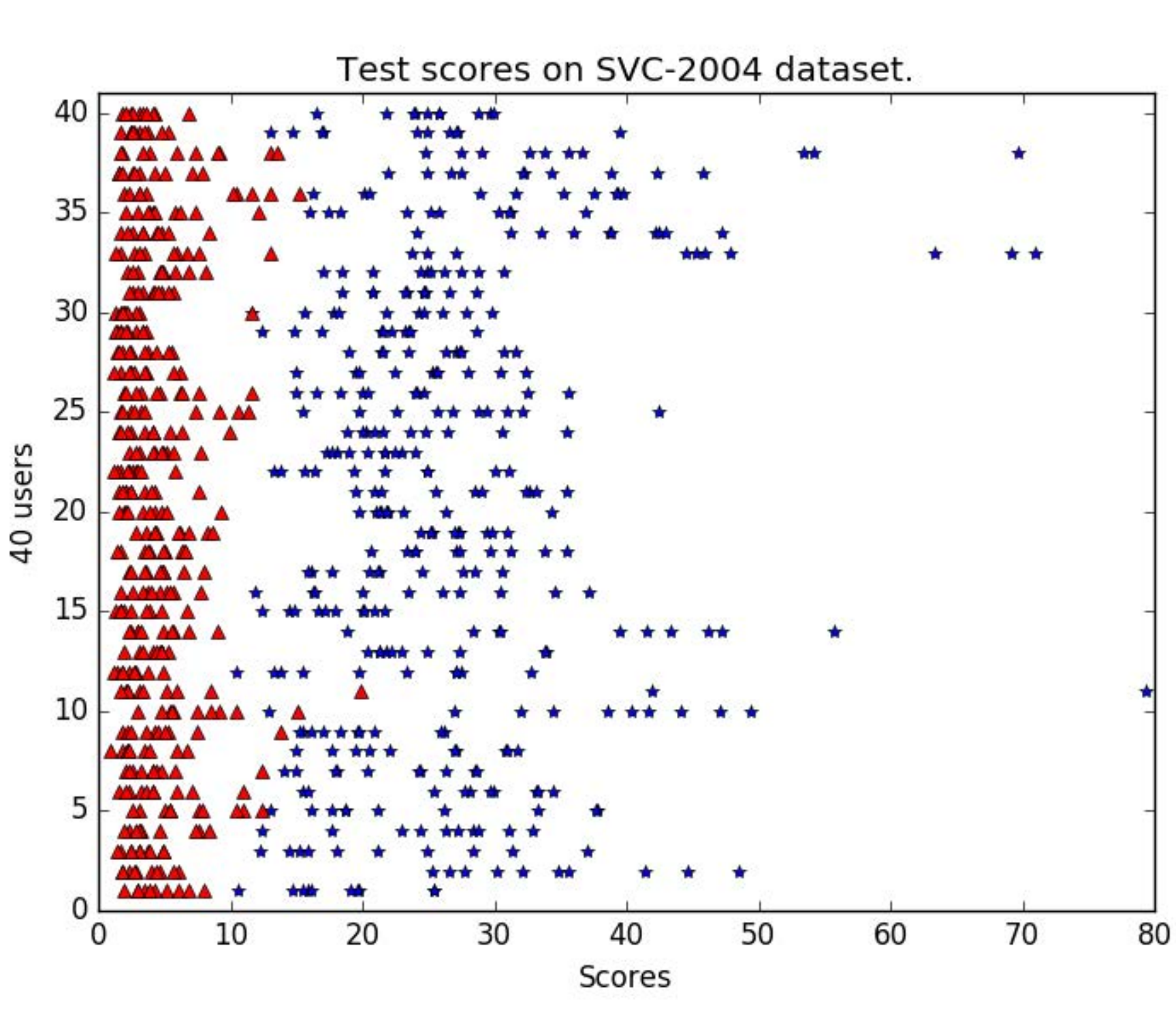}
  \caption{Test scores in one experiment where $W=9$, $m=2$ and $N=10$, on SVC-2004 dataset. The red ones are the genuine signatures while the blue ones are the skilled forgeries. Each row corresponds to one client in SVC-2004 dataset.}\label{test}
\end{figure}

Second, to evaluate the effect of joint training, we set $W=9$, $m=2$, $N=10$ and varied the number of MCYT-100 clients added to the training set. Samples from the first 0, 25, 50 and 75 clients were used for training; the results are given in Table IV, which shows that the signatures from the MCYT-100 dataset help to reduce the EER on the SVC-2004 dataset. The EER is expected to further decrease if more datasets are available. Through joint training, the proposed RNN system can learn more effective representations, whereas the traditional DTW or HMM systems do not have such a property. This property demonstrates the potential for high-accuracy online signature verification when a large amount of data is available.

\begin{table}[tb]
\renewcommand{\arraystretch}{1.2}
\caption{EER (\%) of the proposed RNN system using LNPS descriptor and a varying number of training samples, on SVC-2004 dataset.}
\label{RNNsystem}
\centering
\fontsize{10}{12}\selectfont
\begin{tabular}{ccccccc}
\hline
\multirow{2}{*}{Samples}&\multirow{2}{*}{LNPS}&
\multicolumn{5}{c}{Sliding window size}\cr
\cmidrule(lr){3-7}
&&$W$=7&$W$=9& $W$=11& $W$=13& $W$=15\cr
\midrule
\multirow{4}{*}{N=6}
&$m$=1&6.96&5.79&6.23&7.27&6.22\\
&$m$=2&6.29&5.40&5.84&6.17&6.74\\
&$m$=3&6.36&5.63&5.47&5.64&6.02\\
&$m$=4&6.39&6.27&5.85&5.88&6.89\\\hline
\multirow{4}{*}{N=8}
&$m$=1&2.84&4.01&4.09&5.29&4.57\\
&$m$=2&3.90&3.53&3.84&3.97&5.32\\
&$m$=3&4.61&4.41&4.36&4.55&4.92\\
&$m$=4&4.82&4.29&3.77&4.47&4.43\\\hline
\multirow{4}{*}{N=10}
&$m$=1&2.91&2.44&2.59&3.81&3.53\\
&$m$=2&2.94&2.37&2.56&2.61&3.87\\
&$m$=3&3.58&3.11&3.14&3.09&3.43\\
&$m$=4&3.38&3.12&3.12&3.12&3.17\\
\bottomrule
\end{tabular}
\end{table}

\begin{table}[tb]
\renewcommand{\arraystretch}{1.2}
\caption{EER (\%) of the proposed RNN system using $\triangle x$ and $\triangle y$ as inputs, on SVC-2004 dataset.}
\label{deltaXY}
\centering
\fontsize{10}{12}\selectfont
\begin{tabular}{cccc}
\hline
Samples& $N$=6& $N$=8& $N$=10\\
\midrule
EER&10.92&16.92&9.00\\
\bottomrule
\end{tabular}
\end{table}

To evaluate the system's generalization to unseen clients, we tested on the remaining 25 clients (randomly choosing 10 genuine signatures as templates) when 75 clients from the MCYT-100 dataset were added to the training set in the previous experiment. The test scores are presented in Fig. 5, from which we can see that the scores are tighter and less separable. However, for some clients, e.g., those in the first and 15$^{th}$ rows, the genuine and forgery signatures can be separated without error. As sequential data shows complex dynamic patterns, the datasets we used here could not cover them all. If a large dataset covering different kinds of patterns were available, generalization to unseen clients could be improved. We did not test this supposition due to limited data, and leave it for future work.

\subsection{Comparison with state-of-art methods}
In Table V we compare our results with state-of-art methods on SVC-2004. Note that we use the x-y coordinates only, while some of the methods also uses other time functions, for example, pressure. However, we also use more training data and templates, required by the nature of RNNs.

\begin{figure}
  \centering
  \includegraphics[width=3.5in]{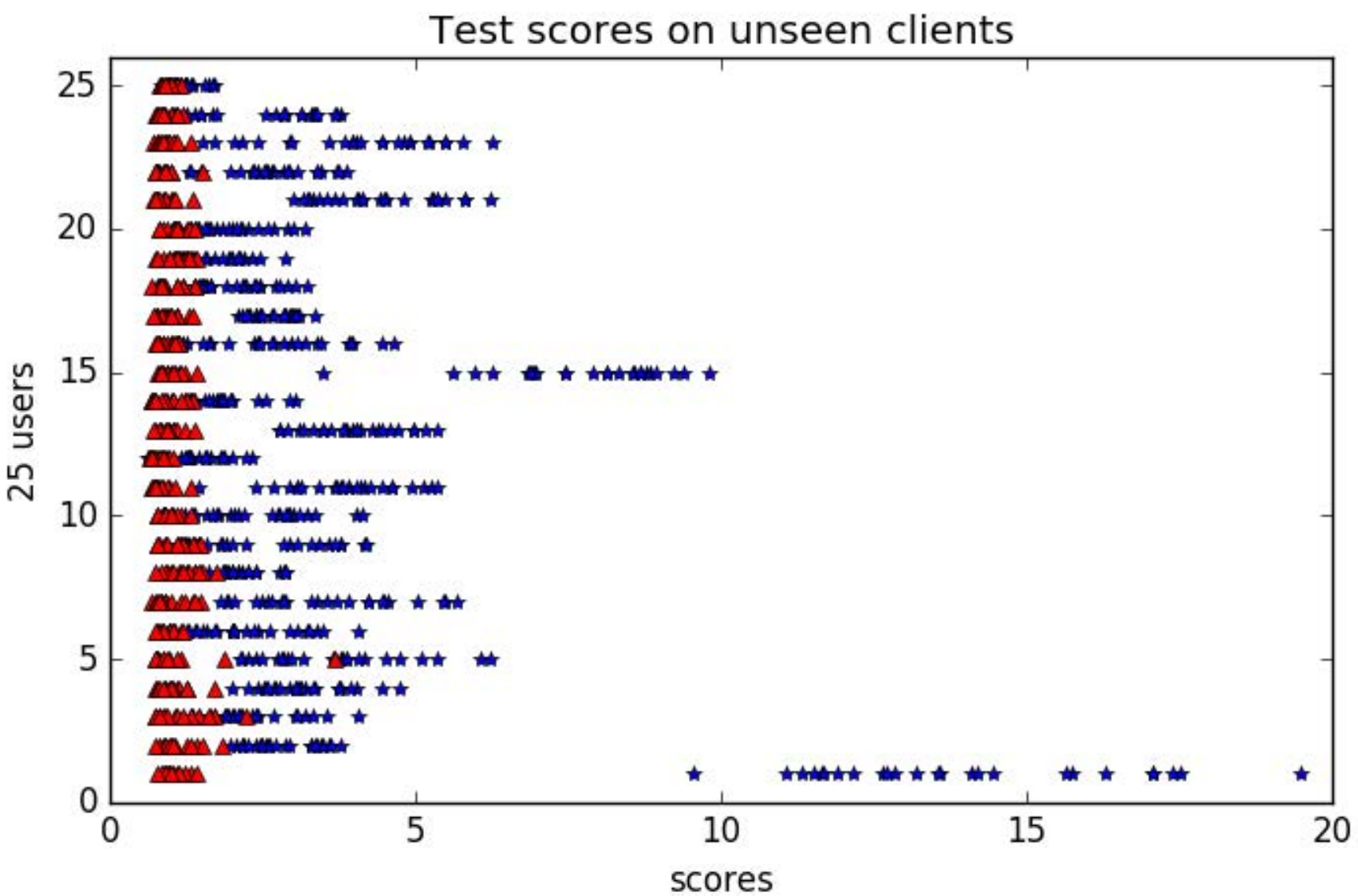}
  \caption{Test scores on unseen 25 clients from MCYT-100. The red ones are the genuine signatures while the blue ones are the skilled forgeries. Each row corresponds to one client.}\label{mcyt25}
\end{figure}

\begin{table}[tb]
\renewcommand{\arraystretch}{1.2}
\caption{EER (\%) of the proposed RNN system using a varying number of MCYT-100 clients for joint training on SVC-2004 dataset.}
\label{user}
\centering
\fontsize{10}{12}\selectfont
\begin{tabular}{cccccc}
\hline
Number of clients&0&25&50&75&100\\\hline
EER (\%)&3.58&3.08&2.79&2.54&2.37\\
\bottomrule
\end{tabular}
\end{table}

\begin{table}[tb]
\renewcommand{\arraystretch}{1.2}
\caption{EER (\%) of the proposed RNN system using LNPS descriptor and a varying number of training samples, on SVC-2004 dataset.}
\label{comparison}
\centering
\fontsize{10}{12}\selectfont
\begin{tabular}{ccc}
Method&EER (\%)&Model\\
\hline
Yeung et al. \cite{yeung2004svc2004}&5.50&DTW \\
Pascual-Gaspar et al. \cite{pascual2009practical} &3.38&DTW \\
Fierrez el at. \cite{Fierrez2007HMM} &6.90&HMM \\
Van et al. \cite{Van2007On} &4.83&HMM \\
Sharma et al. \cite{Sharma2016An} &2.73&DTW\\
Our method &2.37&RNN+LNPS\\
\bottomrule
\end{tabular}
\end{table}

\section{Conclusion}
Inspired by recurrent neural network, we propose a novel RNN system to improve the performance of online signature verification. This study's contributions are twofold. First, a novel descriptor, called the length-normalized path signature (LNPS), is proposed and applied to online signature verification. It has interesting properties and shows promising results. The potential applications of LNPS go beyond online signature verification. For example, LNPS may be applied to online character recognition or text recognition. Second, we propose a novel RNN system for online signature verification that employs metric learning techniques and a joint training scheme. Through joint training, knowledge learned from one dataset can transfer to another dataset. Our proposed RNN system achieves an EER of 2.37\% on the SVC-2004 dataset.

The proposed RNN system also has limitaion. It requires a relatively large training set and heavy computation. It also needs negative samples to generate triplets; however, in some practical cases, only the genuine signatures are available. Therefore, the proposed RNN system needs further investigation and improvement in future work.







%

%
%

\end{document}